\documentclass[runningheads]{llncs}
\usepackage{graphicx}
\usepackage{booktabs}
\usepackage{multirow}
 
\usepackage[margin=1.5in]{geometry}    
\usepackage[english]{babel}
\usepackage[utf8]{inputenc}
\usepackage{algorithm}
\usepackage{arevmath}     
\usepackage[noend]{algpseudocode}

\begin{document}
\title{Enhanced Pub/Sub Communications for Massive IoT Traffic with SARSA Reinforcement Learning \thanks{Work supported by CAPES and Salvador University (UNIFACS)}}
\titlerunning{Enhanced Pub/Sub with SARSA}
%
\author{Carlos E. Arruda\inst{2} \and
Pedro F. Moraes\inst{2} \and
Nazim Agoulmine\inst{1}\orcidID{0000-0002-3461-4284}\and
Joberto S. B. Martins\inst{2}\orcidID{0000-0003-1310-9366}}
\authorrunning{C. Arruda, P. Moraes, N. Agoulmine and J. Martins}
%
\institute{University of Paris-Saclay/IBISC Laboratory, France\and
Salvador University - UNIFACS, Salvador, Brazil\\
\email{arruda.ceas@gmail.com, pmoraes\_@hotmail.com, nazim.agoulmine@univ-evry.fr, joberto.martins@gmail.com}}
\maketitle              
\begin{abstract}
Sensors are being extensively deployed and are expected to expand at significant rates in the coming years. They typically generate a large volume of data on the internet of things (IoT) application areas like smart cities, intelligent traffic systems, smart grid, and e-health. Cloud, edge and fog computing are potential and competitive strategies for collecting, processing, and distributing IoT data. However, cloud, edge, and fog-based solutions need to tackle the distribution of a high volume of IoT data efficiently through constrained and limited resource network infrastructures. This paper addresses the issue of conveying a massive volume of IoT
data through a network with limited communications resources (bandwidth) using a cognitive communications resource allocation based on Reinforcement Learning (RL) with SARSA algorithm. The proposed network infrastructure (PSIoTRL) uses a Publish/ Subscribe architecture to access massive and highly distributed IoT data. It is demonstrated that the PSIoTRL bandwidth allocation for buffer flushing based on SARSA enhances the IoT aggregator buffer occupation and network link utilization. The PSIoTRL dynamically adapts the IoT aggregator traffic flushing according to
the Pub/Sub topic's priority and network constraint requirements.

\end{abstract}

\section{Introduction}\label{intoduction}

Sensors are being extensively deployed and are expected to expand at significant rates in
the coming years in the internet of things (IoT) application areas like smart city, intelligent
traffic systems (ITS), connected and autonomous vehicles (CAV), video analytics in public
safety (VAPS) and mobile e-health \cite{santos_resource_2019} \cite{zhang_openei_2019}, to mention some.

Cloud, edge and fog computing are currently the most prevailing and competing used strategies to convey IoT data between producers and consumers. They use several different alternatives to decide where to process, how to distribute, and where to store IoT data \cite{mukherjee_survey_2018}. Nevertheless, cloud, edge, and fog-based solutions need to tackle the distribution of a high volume of IoT data efficiently.

IoT data distribution, whatever data handling, processing, or storing strategy used, does require efficient network communications infrastructures. In fact, there is a research challenge on enhancing network communications and providing efficient resource allocation based on the fact that, in many applications like IoT, networks have limited and constrained resources \cite{boutaba_comprehensive_2018}.

Machine learning and reinforcement learning with distinct algorithms like Q-Learning
\cite{sutton_introduction_1998}, deep reinforcement learning \cite{koo_deep_2019}, and SARSA (State-Action-Reward-State-Action) \cite{alfakih_task_2020} are being applied to support an efficient allocation of resources in networks for application areas like 5G, cognitive radio, and mobile edge computing, to mention some \cite{cote_using_2018} \cite{rendon_machine_2018}.

The Publish/Subscribe (Pub/Sub) paradigm is extensively used as an enabler for data distribution in various scenarios like information-centric networking, data-centric systems, smart grid, and other general group-based communications \cite{nour_icn_2019} \cite{happ_limitations_2016-1} \cite{an_wide_2015}. The Pub/Sub paradigm is an alternative for IoT data distribution between producers and consumers in general \cite{happ_limitations_2016-1} \cite{moraes_publish/subscribe_2018}.

This work addresses the issue of enhancing the utilization of a communication channel (MPLS label switched path - LSP, physical link, fiber optics slot, others) deployed in net- work infrastructures with limited resources (bandwidth) using the SARSA reinforcement learning algorithm. The target application area is the massive exchange of IoT data using the Pub/Sub paradigm. The framework integrating the SARSA module with the Pub/Sub message exchange deployment (PSIoT described in \cite{moraes_publish/subscribe_2018}) is the PSIoTRL.

Differently from other proposals for network infrastructure resource allocation and optimization that consider, for instance, the quality of service for the entire set of network nodes and Pub/Sub aggregators, the SARSA-based PSIoTRL framework provides a simple data ingress based solution. In fact, it controls the quality of Pub/Sub topics data distribution in an aggregator by keeping Pub/Sub buffer topics occupation below a defined threshold. This indirectly means that a certain amount of bandwidth is available for the Pub/Sub topic and, consequently, a level of quality is allocated for the communication channel. Since the Pub/Sub data transfers are dynamically requested, SARSA deploys a dynamic control of bandwidth allocation for buffer data flushing.

The contribution of this work is multi-fold. Firstly, we modeled the SARSA agent communications with a generic buffer occupation metric. The adopted metric results in a limited amount of SARSA states to allow the allocation of bandwidth preserving Pub/Sub topic priorities without using extensive computational resources. Moreover, we demonstrate that Pub/Sub communications with SARSA can be enhanced by dynamically adjusting IoT aggregators queues occupation to Pub/Sub topics priority and network resources availability.

The remainder of the paper is organized as follows. Section \ref{sec:RelatedWork} presents the related work on IoT data processing deployments with cognitive communications approaches. Section \ref{sec:RL} is a background section about reinforcement learning with SARSA. Sections \ref{sec:PSIoT} and \ref{sec:PSIoTRLArchitecture} describe the basic PSIoT Pub/Sub framework, discuss the RL applicability for the constrained communication problem and present the PSIoTRL framework components including the intelligent orchestrator for IoT traffic management. In Section \ref{sec:ProofOfConcept}, we present a proof of concept for the PSIoTRL and evaluate how it enhances the IoT network resource management. Finally, section \ref{sec:conclusion} concludes with an overview of the main highlights, contributions and future work.

\section{Related Work}\label{sec:RelatedWork}

Architectural and system-wide studies about IoT massive data processing and data flow in smart cities are presented in Al-Fuqaha \cite{mohammadi_enabling_2018-3}, Rathore \cite{rathore_urban_2016} and Martins \cite{martins_towards_2018-1}. Al-Fuqaha \cite{mohammadi_enabling_2018-3} introduces the concept of cognitive smart city where IoT and artificial intelligence are merged in a three-level model with different requirements. In relation to the communication level, the basic approach assumes a fog cloud computing (fog-CC) communication without considering any specific IoT massive traffic requirement. In fact, the proposal assumes that the aggregation with edge processing reduces the required bandwidth for fog-CC communication to a minimum. Rathore \cite{rathore_urban_2016} explores the issue of big data analytics for smart cities. The paper proposes an edge-based aggregation and processing strategy for raw data. Processed data is forwarded through gateways to smart city applications using Internet and assuming bandwidth reduction at edge-level. Martins \cite{martins_towards_2018-1} discusses the potential benefits and impacts of how some technological enablers, like software-defined networking \cite{kreutz_software-defined_2014} and machine learning \cite{xie_survey_2019}, are integrated and aim at cognitive management \cite{rendon_machine_2018}. IoT data edge-processed aggregation, network communication and service deployments towards an efficient overall smart city solution are discussed and the relevance of intelligent communication resource provisioning and deployment are highlighted.

Resource provisioning from the perspective of IoT services deployment for smart city is considered by Santos \cite{santos_resource_2019}. The paper proposes a container-based micro-services approach, using Kubernetes, for service deployment that aims to off-load IoT processing with fog computing. In relation to this paper discussion, the proposal endorses fog processing offloading using a edge-computed approach and does not consider the network communication resource provisioning necessary to distribute the outcomes of the edge processing.

Edge intelligence for service deployment is discussed in Zhang \cite{zhang_openei_2019}. The proposed solutions defines a framework capable to support the deployment of AI algorithms like RL on common edge aggregators like Raspberry. The approach assumed is to enable the execution of AI algorithms in the edge for those applications that require near real time edge processing like voice recognition and on-board autonomous vehicles processing.

Machine learning in communication networks is broadly addressed by Cote \cite{cote_using_2018} and Boutaba et al in \cite{boutaba_comprehensive_2018}.

Intelligent network communication resources are considered by Zhao  \cite{zhao_routing_2019-1}. The work presents a smart routing solution for crowdsourcing data with mobile edge computing (MEC) in smart cities using reinforcement learning (RL). The solution defines routes, differently from ours that optimizes the bandwidth allocated for IoT flow flushing.

In summary, this work advances on existing studies that propose service provisioning by adding an intelligent component for the allocation of communication resources between IoT data aggregators and IoT data consumers. From the architectural point of view, this work adopts an edge-based processing approach coupled with an efficient communication resource allocation for massive IoT data transfers.

\section{Reinforcement Learning and SARSA Algorithm} \label{sec:RL}

Reinforcement learning (RL) is a largely used machine learning (ML) technique in which a trial-and-error learning process is executed by an agent that acts on a system aiming to maximize its rewards. The RL algorithm is expected to learn how to reach or approach a certain objective by interacting with a system through a feedback loop \cite{latah_artificial_2019-1} \cite{sutton_introduction_1998} \cite{moerland_framework_2020}.

In RL, a reward value \textit{r} is received by the agent for the transitions from one state to another. The overall objective of the agent is to find a policy $\pi$ which maximizes the expect future sum of rewards received, each of them, subjected to a discount policy $\gamma$.

The \textit{value function} in RL is a prediction of the return available from each state, as indicated in Equation \ref{eq:ValueFunction1}.

\begin{eqnarray}\label{eq:ValueFunction1}
V(x_t) \leftarrow E\{\sum_{k=o}^{\infty}\gamma^{k}r_{t+k}\}
\end{eqnarray}

Where $r_t$ is the reward received for the transition from state ${x_t}$ to ${x_{t+1}}$ and $\gamma$ is the discount factor $(0 \leq \gamma \leq 1)$. The value function  $V(x_t)$ represents the discounted sum of rewards received from step \textit{t} onward. Therefore, it will depend on the sequence of actions taken and on the policy adopted to take these actions.

Two well-known and somehow similar reinforcement learning algorithms are Q-learning \cite{sutton_introduction_1998} \cite{dabbaghjamanesh_reinforcement_2020} and SARSA \cite{liao_dynamic_2020} \cite{alfakih_task_2020}.

Q-learning is an off-policy RL algorithm in which the agent finds an optimal policy that maximizes the total discount expected reward for executing a particular action at a particular state. Fundamentally, Q-learning finds the optimal policy in a step-by-step manner. Q-learning is off-policy because the next state and action are uncertain when the algorithm updates the value function.

In Q-learning, the value function termed Q-function is learnt. It is a prediction of the return associated with each action $a \in A$ (set of actions). This prediction can be updated with respect to the predicted return of the next state visited (Equation \ref{eq:PredictNextQ}).

\begin{eqnarray}\label{eq:PredictNextQ}
Q(x_t, \textit{a}_t) \leftarrow \textit{r}_t + \gamma V(x_{t+1})
\end{eqnarray}

Since the overall objective is to maximize the reward received, the current estimate of $V(x_t)$ becomes:

\begin{eqnarray}\label{eq:PredictNextQ1}
Q(x_t, \textit{a}_t) \leftarrow \textit{r}_t + \gamma \max_{a \in A} Q(x_{t+1},a)
\end{eqnarray}

The Q-function is shown to converge for markovian decision processes (MDP) \cite{ramani_short_2019}. In Q-learning, the agent maintains a lookup table of $Q(X,A)$ and $Q(x,a)$  represents the current estimate of the optimal action value function. Once the Q-function has converged, the optimal policy $\pi$ is to take the action in each state with the highest predicted return (greedy policy) \cite{rummery:cuedtr94}.

\subsection{SARSA Algorithm}

SARSA (State-Action-Reward-State-Action) algorithm is the reinforcement learning approach used by PSIoTRL framework \cite{liao_dynamic_2020} \cite{alfakih_task_2020}.

SARSA is a temporal-difference (TD) on-policy algorithm that learns the Q-values based
on the action performed by the current policy. SARSA algorithm differs from Q-learning
by the way it sets up the future reward. In SARSA the agent uses the action and the state
at time $t + 1$ to update the Q-value. The SARSA tuple of main elements involved in the
interaction process are:

\begin{eqnarray}\label{eq:PredictNextQ2}
<x_t, a_t, r_{t+1}, x_{t+1},a_{t+1}>
\end{eqnarray}

Where:
\begin{itemize}
    \item $x_t, a_t$ are the current state and action;
    \item $r_{t+1}$ is the reward; and
    \item $x_{t+1},a_{t+1}$ are the next state and action reached using the policy ($\epsilon-Greedy$).
\end{itemize}

SARSA Q-values are therefore updated based on the Equations \ref{eq:SARSA} or \ref{eq:SARSADelta}.

\begin{eqnarray}\label{eq:SARSA}
Q(x_t,a_t) \leftarrow Q(x_t,a_t) + \alpha[r_{t+1} + \gamma Q(x_{t+1},a_{t+1}) - Q(x_t, a_t)]
\end{eqnarray}

\begin{eqnarray}\label{eq:SARSADelta}
\Delta Q(x_{t-1},a_{t-1}) = \alpha (t) [r_{t-1} + \gamma Q(x_t,a_t) - Q(x_{t-1}, a_{t-1})]
\end{eqnarray}

Where:
\begin{itemize}
    \item $\alpha$ is the learning rate $(0 \leq\alpha\leq1)$; and
    \item $\gamma$ is the discount factor $(0 \leq \gamma \leq 1)$.
\end{itemize}

\subsection{Network Communications Resource Allocation with SARSA}

Reinforcement learning has achieved superhuman performance in games like Go \cite{silver_mastering_2017} and also obtained a critical result bridging the divide between high-dimensional sensory inputs and actions with ATARI \cite{mnih_human-level_2015}.

Video games and communication networks have imperfect but interesting operation similarities. As discussed in Cote \cite{cote_using_2018}, video games and networks are closed systems with a finite number of states. In games, actions include pressing buttons and moving the joystick, the image pixels define the state, and the cost function is the game score. In networks, actions correspond mainly to network configuration parameters (bandwidth allocation, fiber allocation, others) or network routing configuration. The network state (snapshot) defines the state of the system at each iteration, and the cost function corresponds to the performance of the network such like utilization rate, throughput, or number of dropped packets.

Hence, it is reasonable to consider that SARSA may have a parallel success for allocating
network communications resources (bandwidth) in the PSIoTRL.

Reinforcement learning with SARSA has proved efficient and obtained substantial success in resource allocation \cite{alfakih_task_2020} \cite{sampaio_using_2018}, cloud computing \cite{bibal_benifa_rlpas_2019} and computational offloading \cite{asghari_task_2020} \cite{nassar_reinforcement_2019}.

An evaluation of SARSA and various other model-free algorithms is presented in Dafazio
\cite{defazio_comparison_2014}. The evaluation considered diverse and difficult problems within a consistent environment (Arcade \cite{machado_revisiting_2018}) involving configuration aspects like Epsilon-greedy policy, exploration versus exploitation and state space with SARSA outperforming algorithms like Q-learning, ETTR (Expected Time-to-Rendezvous) \cite{wang_reinforcement_2019}, R-learning \cite{mahadevan_average_1996}, GQ algorithm \cite{wang_finite-sample_2020} and Actor-Critic algorithm \cite{ciosek_better_2019}.

SARSA algorithm features and potential advantages within the PSIoTRL environment include:
\begin{itemize}
    \item Being an on-policy algorithm, SARSA attempts to evaluate or improve the policy that
is used to make decision;
    \item SARSA avoids the state explosion issue common with the Q-learning algorithm;
    \item PSIoTRL has a reduced state space and, consequently, SARSA behaves effectively by
exploring all states at least once; and
    \item Key issues that RL algorithms experience like bad initial performance and large training
time are minimized in the PSIoTRL environment since SARSA addresses the allocation of network bandwidth locally with a reduced state-space.
\end{itemize}

\section{The Publish/Subscribe Framework for IoT (PSIoT)}\label{sec:PSIoT}

\begin{figure}
\centerline{\includegraphics[width=220pt,height=15.3pc]{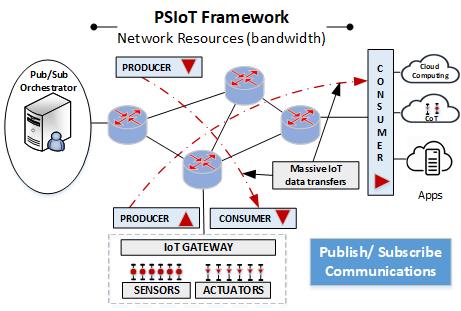}}
\caption{PSIoT Framework Functional View \cite{moraes_publish/subscribe_2018}.\label{fig:PSIoT}}
\end{figure}

The objective of this work is to enhance the Publish/ Subscribe (Pub/Sub) communications used in the context of massive IoT data transfers (Figure \ref{fig:PSIoT}).

The Pub/Sub communications is supported by the PSIoT framework described in \cite{moraes_publish/subscribe_2018}. The PSIoT framework aims to efficiently handle network resources and IoT QoS requirements over the network between the IoT devices and consumer IoT applications \cite{moraes_publish/subscribe_2018}. Consumers are applications executed in servers located beyond the backbone network, cloud computing infrastructure accessed by the network or any other scheme that makes use of the managed network infrastructure for communications (Figure \ref{fig:PSIoT}).

The PSIoT framework was developed to manage IoT traffic in a network and to provide QoS based on IoT data characteristics and network-wide specifications, e.g. total network use, realtime network traffic, routing and bandwidth constraints.

IoT data generated from sensors and devices can be aggregated and processed in the cloud or at the network edge, where each of these points are considered as \textit{Aggregators} or \textit{Producers} in the PSIoT framework. Whereas each client that receives IoT data, both end-users applications and even other aggregation nodes, are denoted as \textit{Consumers}.

The PSIoT framework was designed to orchestrate massive IoT traffic and allocate network resources between producers and consumers. Besides, the PSIoT framework can schedule the data flow, based on Quality of Service (QoS) requirements, and allocates bandwidth on a backbone with limited resources. For this, the PSIoT was modeled to operate with four main components: producers, consumers, a backbone with limited resources; and the orchestrator (Figure \ref{fig:PSIoT}) \cite{moraes_publish/subscribe_2018}.
 
Aggregators are elements that gather the data obtained by connected sensors to send them (in an opportune moment) to their consumers (clients) using a Pub/Sub-style communication channel. In the opposite direction, aggregators deliver data received from producers to send them to actuators.
 
This exchange of information is performed through the Pub/Sub architecture, whose characteristic of asynchronous, use of topics and the use of a broker, makes it attractive for IoT applications \cite{happ_limitations_2016-1}.
 
PSIoT implements QoS when forwarding the data gathered by the aggregator to output buffers, according to the following characteristics \cite{moraes_publish/subscribe_2018}:

 \begin{itemize}
\item b0 (priority) - high transmission rates and low delay application requirements for health care and data from critical industrial sensors, as examples;
\item b1 (sensitive) - commercial data and security sensors; and
\item b2 (insensitive) - best effort.
\end{itemize}

\begin{figure}[t]
\centerline{\includegraphics[width=260pt,height=17.6pc]{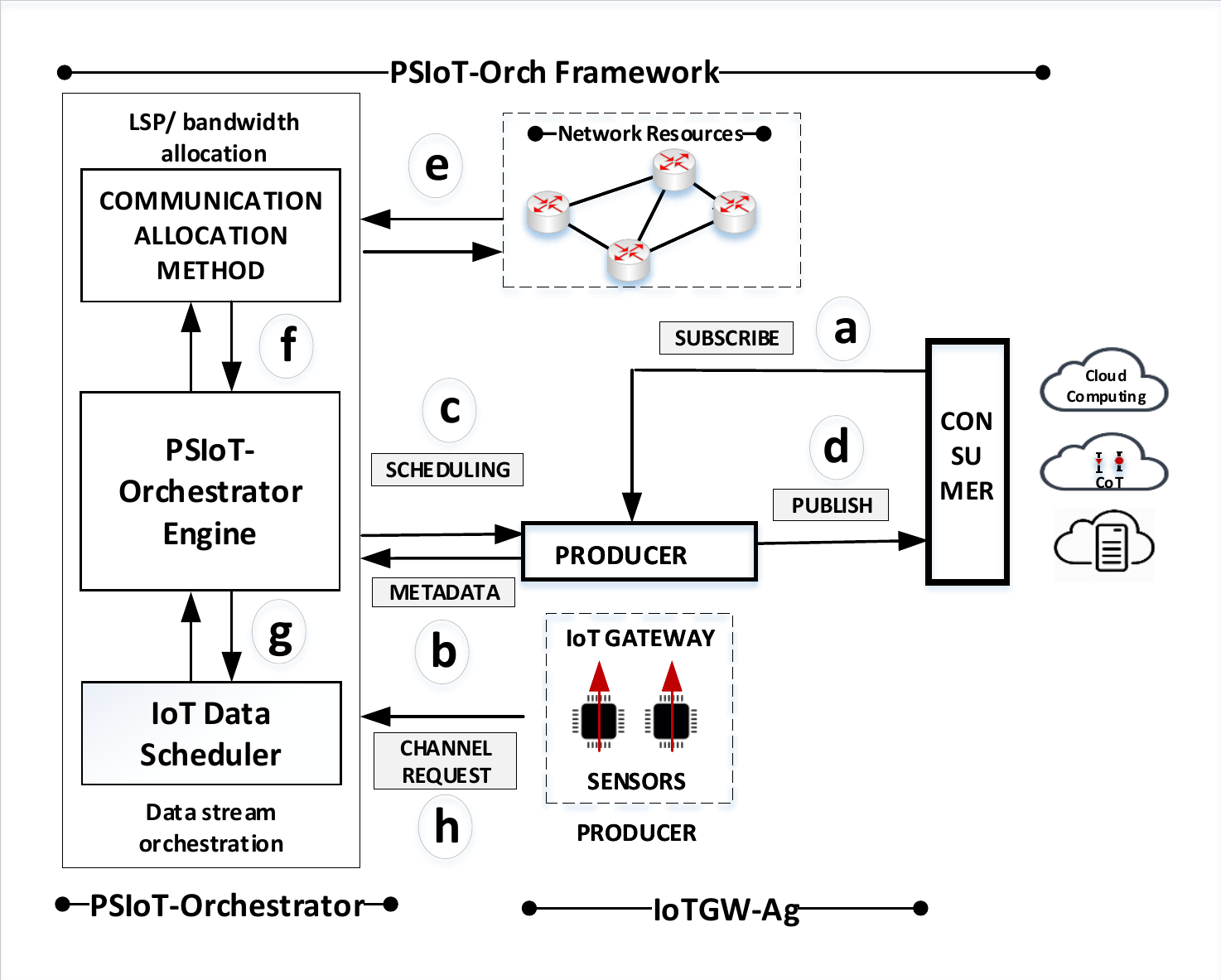}}
\caption{PSIoT Pub/Sub Message Model \cite{moraes_publish/subscribe_2018}.\label{fig:PubSubMessages}}
\end{figure}

The Pub/Sub message model adopted by the PSIoT is illustrated in Figure \ref{fig:PubSubMessages}. In summary, the message flow is as follows \cite{moraes_publish/subscribe_2018}:

\begin{enumerate}
    \item A consumer subscribes to a particular topic and specify the requested QoS level with a particular aggregator ($a$);
    \item The aggregator sends to the orchestrator relevant metadata such as number of subscribers and their associated QoS levels and buffer allocation ($b$);
    \item The orchestrator notifies the aggregator with an amount of bandwidth that can be consumed by each level of QoS ($c$); and
    \item The aggregator publishes the data to the consumer according to bandwidth and data availability in the buffer ($d$).
\end{enumerate}

The PSIoT uses fixed rule scheduling for IoT data consumption.

\section{PSIoTRL Framework - Architectural Components, Communication Model and SARSA Algorithm}\label{sec:PSIoTRLArchitecture}

\begin{figure}[t]
\centerline{\includegraphics[width=300pt,height=250pt]{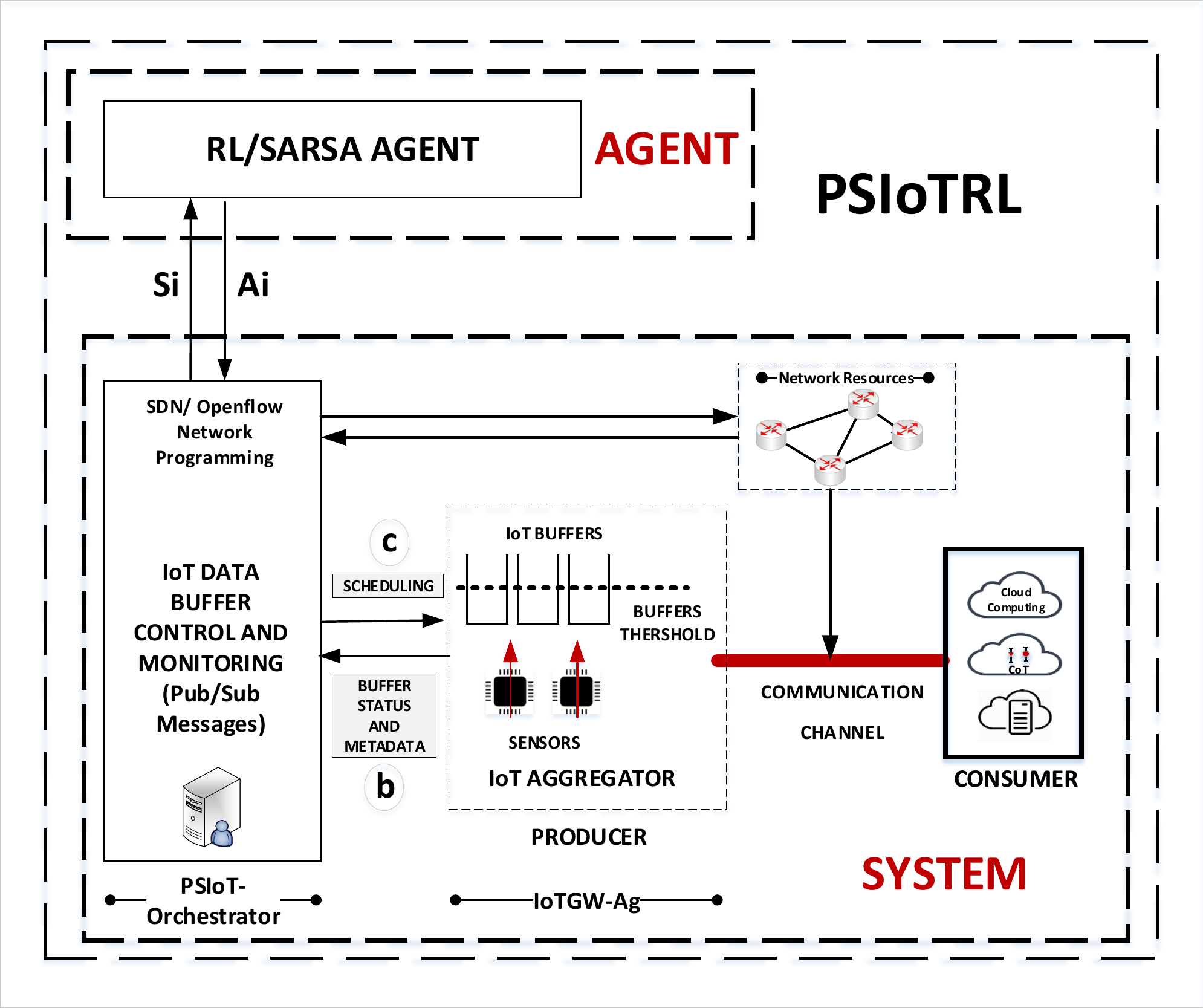}}
\caption{PSIoTRL Components and Communication - Orchestrator, Aggregator, Buffers and Communication Chanel.
\label{fig:PSIoTRL-Communication}}
\end{figure}

The PSIoTRL architectural components are illustrated in Figure \ref{fig:PSIoTRL-Communication}. Its main proposed modules are:

\begin{itemize}
    \item The SARSA agent;
    \item The PSIoTRL orchestrator module;
    \item Aggregators (at least one for each cluster where IoT data will be consumed);
    \item The network infrastructure (backbone); and
    \item Producers and consumers.
\end{itemize}

The SARSA agent is integrated in the PSIoT framework \cite{moraes_publish/subscribe_2018}  and, by demand, allocates bandwidth for the aggregator queues.

The PSIoTRL orchestrator module is described in \cite{moraes_publish/subscribe_2018} and \cite{moraes_pub/sub_2019}. It basically has the knowledge of each aggregator Pub/Sub subscriptions, as well as the QoS levels required for each topic subscription. This allows it to control the transmission emptying rates of IoT data from each buffer within the aggregator in the network. The SARSA agent computes the allocated bandwidth and the orchestrator deploys it.

The aggregators are Fog-like nodes connected to IoT devices that act as aggregators for their data and also act as Pub/Sub producers regarding IoT applications subscribing to topics and consuming corresponding IoT data \cite{moraes_pub/sub_2019} \cite{happ_limitations_2016-1}.

The network interconnects data producers (aggregators) and consumers and, has limited
bandwidth resources for massive IoT data transfers.

Producers and consumers exchange IoT data. They use Pub/Sub [10] to communicate
and exchange massive amounts of IoT data through a network with constrained bandwidth
resources \cite{moraes_publish/subscribe_2018} \cite{moraes_pub/sub_2019}.

\subsection{PSIoTRL Aggregator Configuration}

The PSIoTRL framework uses the Pub/Sub paradigm to produce and consume data. Consumers request IoT data on the aggregator's queues, and the aggregator empties its buffers according to consumer requests. The aggregator queues are deployed with the following configuration:

\begin{itemize}
    \item Three IoT data queues (buffers) are configured per aggregator (one Pub/Sub topic per queue): $B1$, $B2$ and $B3$;
    \item Initial buffer transmission rates\footnote{By initial buffer transmission rate, we mean the configured initial transmission rate to empty buffers without any buffer over-utilization.} are: $T1$, $T2$ and $T3$; and
    \item Buffer priorities are: $p1$, $p2$ and $p3$ with $p1>p2>p3$.
\end{itemize}

For this evaluation we consider one aggregator $Ag_i$ located at network node $n_z$ that
delivers IoT data to a set of consumers $C_i$, with $i = 1, 2, ..., n$ located at network node $n_y$. Between nodes $n_z$ and $n_y$ there is a communication resource (MPLS LSP, physical link, fiber slot, other) with a limited bandwidth $BW_{nz}$. The communication resource interconnecting
Pub/Sub producer and consumers is then constrained as follows:

\begin{eqnarray}\label{eq:BandwidthLimit}
BW_{zy} \ge \sum_{j=1}^{n}TB_{j}\
\end{eqnarray}

Where $TB_{j}$ is the currently transmission rate to empty a buffer $j$.

The aggregator monitors buffer occupation and occupation above a defined threshold limit is signaled to the orchestrator (Figure \ref{fig:PSIoTRL-Communication}).

\subsection{The SARSA Agent Communication Model} \label{sec:PSIOTRLCOM}

\begin{center}
\begin{table}[t]%
\centering
\caption{Buffer (Queue) States.\label{TableStates}}%
\begin{tabular*}{330pt}{@{\extracolsep\fill}lcccc@{\extracolsep\fill}}
\toprule
\textbf{Buffer (queue) State (B1,B2,B3)} & \textbf{Description} \\
\midrule
BL, BL, BL & all queues below the threshold limit  \\ 
BL, BL, AL & queues 1 and 2 below the threshold limit \\ 
BL, AL, BL & queues 1 an 3 below the threshold limit \\ 
BL, AL, AL & queue 1 below the threshold limit \\ 
AL, BL, BL & queues 2 and 3 below the threshold limit \\ 
AL, BL, AL & queue 2 below the threshold limit \\ 
AL, AL, BL & queue 1 below the threshold limit \\ 
AL, AL, AL & no queues below the threshold limit \\ 
\bottomrule
\end{tabular*}
\item[$\dagger$] Legend: BL - Below Limit; AL - Above Limit.
\end{table}
\end{center}

The PSIoTRL deployment and operation require SARSA agent modeling for the buffer bandwidth allocation problem and the definition of basic SARSA algorithm configuration parameters.

The primary goal of the SARSA agent is to arbitrate the output transmission rates among the aggregator buffers in order to efficiently distribute IoT data to the consumers (Figure \ref{fig:PSIoTRL-Communication}).

The PSIoTRL SARSA agent is modeled with a finite set of states for the aggregator output queues, a finite set of configuration actions to be executed on each queue, and a set of reward values for each state/action transition pair.

In the SARSA agent, the system state \textit{\(S_i\)} represents the overall aggregator's output queue status in terms of occupation at a given time \textit{\(t_i\)}.

The SARSA agent states are indicated in Table \ref{TableStates}. Each queue has two states either below (BL) or above (AL), a preconfigured threshold. Having two states for each queue allows, in this context, sufficient information for bandwidth allocation and contributes to avoid the common state explosion issue existing in many RL deployments.

A discrete and small number of states for the queues can be reasonably assumed because it mainly works as a threshold to indicate that queues require attention to enforce priorities or maximize throughput. The first case eventually happens when queued Pub/Sub messages require more capacity to be transferred than the actually allocated one. The second case corresponds to the situation where some queues have plenty of data to transmit, while others have unused capacity.

The utilization of the buffer threshold results in having a kind of agnostic Pub/Sub implementation. In fact, the threshold-based cognitive actions allow:
\begin{itemize}
    \item To tune the PSIoTRL to have a faster or proactive reaction to adjust the queue's bandwidth; and
    \item To tune the Pub/Sub dynamic reconfiguration capability according to IoT data transfer sensitivity.
\end{itemize} 

The PSIoTRL SARSA agent actions are illustrated in Table \ref{tab:PSIoTRLActions}. Three actions are defined for each buffer: increase capacity (transmission rate), reduce capacity (transmission rate) and do nothing. Therefore, twenty seven actions are possible for each agent state.  The amount of bandwidth increased or reduced per queue is a SARSA configuration parameter. A set of rewards is also defined for each executed state/action pair.

\begin{center}
\begin{table}[t]%
\centering
\caption{SARSA agent actions.\label{tab:PSIoTRLActions}}%
\begin{tabular*}{170pt}{@{\extracolsep\fill}lcccc@{\extracolsep\fill}}
\toprule
\textbf{Buffer (Queue)} & \textbf{Actions} \\ 
\midrule
B1 & $T+$, $T-$, N  \\ 
B2 & $T+$, $T-$, N \\ 
B3 & $T+$, $T-$, N \\ 
\bottomrule
\end{tabular*}
\item $T+$: increase~transmission~rate;~$T-$:~decrease~transmission~rate;~N:~null~action.
\end{table}
\end{center}

\subsection{The SARSA $\epsilon$-Greedy Policy}

The SARSA agent uses an $\epsilon$-greedy policy  since it must exploit as much as possible the
acquired knowledge but also explores new possibilities of enhancing the allocation of band-
width for queue communications. The SARSA agent will take new actions at random with
the  $\epsilon$-greedy policy defining the probability of choosing random actions. No value change
or on-the-fly fine-tuning of \(\epsilon\) was considered in this solution.

The \(\epsilon\)-greedy policy matches adequately the inherent need of a Pub/Sub IoT message
delivery framework. Random demands from consumers consume Pub/Sub data with different priorities. These demands generate a random volume of data to be transferred from aggregator queues to consumers. The per-queue bandwidth distribution must be dynamically computed among IoT data queues considering the existing constraint (bandwidth limitation) of the available communication channel.

\subsection{The SARSA Algorithm for Bandwidth Allocation}

The SARSA agent manages the aggregator's output queues transmission rate according to IoT data demands, IoT data priority, and communication resource constraints (bandwidth limitation). The bandwidth allocation algorithm pseudo-code is presented in Algorithm \ref{alg:PSIoTAigorithm}.

\begin{algorithm}
\caption{PSIoTRL-SARSA Bandwidth Allocation Algorithm Pseudo-code}\label{alg:PSIoTAigorithm}

\begin{algorithmic}[1]
\Procedure{PSIoTRL-SARSA}{Q({$S_t,A_t,r, S_{t+1},A_{t+1}$})}\Comment{SARSA states and reward}
    \For {each pair $(Q_t,A_t)$}\Comment{Initialization}
    \State Initialize Q-values - \(Q(S_t,A_t)=0\) \Comment{\textit{Tabula rasa} approach}
    \EndFor
    \Repeat \Comment{Forever}
    \State Gets current PSIoTRL buffers state $S_t$ \Comment{Buffer state is the trigger event}
    \Repeat \Comment{Finishes upon terminal condition}
        \State Choose action $A_t$ using the \(\epsilon\)-greedy policy
                \Comment{Exploration and exploitation}
        \State Execute action $A_t$ on the PSIoTRL system
                \Comment{T+, T- or do nothing on queues}
        \State Get immediate reward $r_t$
        \State Collect new state $S_{t+1}$
        \State Choose new action $A_{t+1}$ using the \(\epsilon\)-greedy policy
        \State Update Q-value - $Q(S_t,A_t)$ using SARSA equation \ref{eq:SARSA}
                \Comment{SARSA algorithm}
    \Until{Terminal condition reached}
    \Until{forever}
\EndProcedure
\end{algorithmic}
\end{algorithm}

The algorithm procedure is as follows:
\begin{enumerate}
    \item The algorithm starts initializing the Q-values in table, $Q(s, a)$;
    \item The current state, $s$ is captured;
    \item An action $a$ is chosen for the current state using the greedy policy;
    \item The agent triggers the action, and observes the immediate reward, $r$, as well as the new reached state $s'$;
    \item The Q-value for the state $s'$ is updated using the observed reward and the maximum reward possible for the next state; and
    \item Finally for the current cycle, set the state to the new state, and repeat the process until the end condition is reached.
\end{enumerate}

The end condition for the PSIoTRL bandwidth allocation process is the following:
\begin{itemize}
    \item Bandwidth constraint limit is reached (Equation \ref{eq:BandwidthLimit}); or
    \item Current buffers transmission rates corresponding  priorities are ($T1>T2>T3)$; or
    \item Maximum number of attempts is reached.
\end{itemize}

\section{Proof of Concept} \label{sec:ProofOfConcept}

The purpose of this proof of concept is to validate that the agent behavior enhanced the allocation of bandwidth when one or more buffers exceeded the defined buffer occupation limit (50\% - AL condition). When this event occurs, the aggregator detects the problem and sends an alarm containing metadata to the orchestrator. After that, the SARSA agent allocates a new percentage of bandwidth among the aggregator buffer.

Four initial buffer conditions were used:
\begin{itemize}
    \item One queue exceeds the occupation threshold $(AL,BL,BL)$;
    \item Two queues exceed the occupation threshold $(AL,AL,BL)$; 
    \item Three queues exceed the occupation threshold $(AL,AL,AL)$; and
    \item One queue exceeds the link bandwidth capacity and the two other queues exceed the occupation threshold $(+AL,AL,AL)$.
\end{itemize}

The considered performance parameters are the following:

\begin{itemize}
        \item Queue (buffer) Occupation;
        \item Link Occupation; and
        \item Packet Loss.
\end{itemize}

The allocation of bandwidth to the aggregator queues is evaluated in two scenarios. In scenario 1, a predefined simple algorithm is used to allocate the bandwidth. In scenario 2, the SARSA agent does the same task. Table \ref{TableProofofConcept} presents a summary of the evaluation scenarios.

\begin{table}[!htp]
\centering
\caption{Summary of the Proof of Concept Scenarios.
\label{TableProofofConcept}}%
\begin{tabular}{@{}lllllllc@{}}
\toprule
\multicolumn{4}{c}{\textbf{AGGREGATOR}} & \textbf{} & \textbf{} & \multicolumn{2}{c}{\textbf{ORCHESTRATOR}} \\ \midrule
\textbf{Queues States (B1,B2,B3)} & \multicolumn{1}{c}{\textbf{T1i}} & \multicolumn{1}{c}{\textbf{T2i}} & \multicolumn{1}{c}{\textbf{T3i}} & \multicolumn{1}{c}{\textbf{}} & \multicolumn{1}{c}{\textbf{}} & \multicolumn{1}{c}{\textbf{Scenario 1}} & \textbf{Scenario 2 (SARSA)} \\
AL,BL,BL &  &  &  &  &  &  & \multicolumn{1}{l}{} \\
AL,AL,BL & 35\% & 25\% & 15\% &  &  & T1 = T1i * factor; & SARSA \\
AL,AL,AL &  &  &  &  &  & T2 = (100 - T1)/2; & Module \\
$+AL$,BL,BL &  &  &  &  &  & T3 = (100 - T1)/2; & \multicolumn{1}{l}{} \\ \bottomrule
\end{tabular}
\begin{itemize}
\item $+AL$: 120\%~buffer~occupation;~dropping~packets;~factor:1.15~or~1.25~or~1.50.
\end{itemize}
\end{table}

\subsection{Scenario 1 - Bandwidth Allocation with Fixed Rule}

In this scenario, the orchestrator has a simple fixed rule to increase or decrease the bandwidth for buffer flushing. $T1$ bandwidth is adjusted by a fixed factor (15\%, 25\% or 50\%) and the remaining queues get half of the remaining bandwidth as follows:
\begin{itemize}
    \item $T1=T1_i*$factor where $T1_i=$ $T1$ initial state;
    \item $T2=(100-T1)/2$; and
    \item $T3=(100-T1)/2$.
\end{itemize}

Buffer transmission emptying initial rates are respectively 35\%, 25\% and 15\% for T1, T2 and T3. Tables \ref{Result15Case1}, \ref{Result25Case1} and \ref{Result50Case1} present the results obtained with the fixed rule bandwidth allocation method. In these Tables, packets loss for queue $B_i$ is $P_i$.

\begin{table}[!htp]
\centering
\caption{Bandwidth allocation with fixed rule - 15$\%$ bandwidth adjustment.\label{Result15Case1}}
\resizebox{\textwidth}{!}{%
\tiny{
\begin{tabular}{@{}ccccc@{}}
\toprule
\textbf{\begin{tabular}[c]{@{}c@{}}Initial QS \\ (B1i,B2i,B3i)\end{tabular}} & \textbf{\begin{tabular}[c]{@{}c@{}} Final QS\\ (B1,B2,B3)\end{tabular}} & \textbf{\begin{tabular}[c]{@{}c@{}} Final TR\\ (T1,T2,T3)\end{tabular}} & \textbf{\begin{tabular}[c]{@{}c@{}} Packet Loss \\ (P1,P2,P3)\end{tabular}} & \textbf{ Link Occupation} \\ \midrule
AL,BL,BL & (72,16,0) & (40,30,30) & (0,0,0) & 100 \\
AL,AL,BL & (72,64,0) & (40,30,30) & (0,0,0) & 100 \\
AL,AL,AL & (72,64,0) & (40,30,30) & (0,0,0) & 100 \\
+AL,BL,BL & (108,16,0) & (40,30,30) & (8,0,0) & 100 \\ \bottomrule
\end{tabular}%
}}
\begin{itemize}
\item QS: Queue~State; TR: Transmission~Rate; AL: 80$\%$; BL: 20$\%$; +AL: 120$\%$.
\end{itemize}
\end{table}

\begin{table}[!htp]
\centering
\caption{Bandwidth allocation with fixed rule - 25$\%$ bandwidth adjustment.
\label{Result25Case1}}
\resizebox{\textwidth}{!}{%
\tiny{
\begin{tabular}{@{}ccccc@{}}
\toprule
\textbf{\begin{tabular}[c]{@{}c@{}}Initial QS \\ (B1i,B2i,B3i)\end{tabular}} & \textbf{\begin{tabular}[c]{@{}c@{}} Final QS\\ (B1,B2,B3)\end{tabular}} & \textbf{\begin{tabular}[c]{@{}c@{}} Final TR\\ (T1,T2,T3)\end{tabular}} & \textbf{\begin{tabular}[c]{@{}c@{}} Packet Loss \\ (P1,P2,P3)\end{tabular}} & \textbf{Link Occupation} \\ \midrule
AL,BL,BL & (56,18,2) & (44,28,28) & (0,0,0) & 100 \\
AL,AL,BL & (56,72,2) & (44,28,28) & (0,0,0) & 100 \\
AL,AL,AL & (56,72,8) & (44,28,28) & (0,0,0) & 100 \\
+AL,BL,BL & (84,18,2) & (44,28,28) & (0,0,0) & 100 \\ \bottomrule
\end{tabular}%
}}
\begin{itemize}
\item QS: Queue~State; TR: Transmission~Rate; AL: 80$\%$; BL: 20$\%$; +AL: 120$\%$.
\end{itemize}
\end{table}

\begin{table}[!htp]
\centering
\caption{Bandwidth allocation with fixed rule - 50$\%$ bandwidth adjustment.\label{Result50Case1}}

\resizebox{\textwidth}{!}{%
\tiny{
\begin{tabular}{@{}ccccc@{}}
\toprule
\textbf{\begin{tabular}[c]{@{}c@{}}Initial QS \\ (B1i,B2i,B3i)\end{tabular}} & \textbf{\begin{tabular}[c]{@{}c@{}} Final QS\\ (B1,B2,B3)\end{tabular}} & \textbf{\begin{tabular}[c]{@{}c@{}} Final TR\\ (T1,T2,T3)\end{tabular}} & \textbf{\begin{tabular}[c]{@{}c@{}} Packet Loss \\ (P1,P2,P3)\end{tabular}} & \textbf{Link Occupation} \\ \midrule
AL,BL,BL & (40,20,10) & (53,24,23) & (0,0,0) & 100 \\
AL,AL,BL & (40,80,10) & (53,24,23) & (0,0,0) & 100 \\
AL,AL,AL & (40,80,40) & (53,24,23) & (0,0,0) & 100 \\
+AL,BL,BL & (60,20,10) & (53,24,23) & (0,0,0) & 100 \\ \bottomrule
\end{tabular}%
}}
\begin{itemize}
\item Legend: QS: Queue~State; TR: Transmission~Rate; AL: 80$\%$; BL: 20$\%$; +AL: 120$\%$.
\end{itemize}

\end{table}

\subsection{Scenario 2 - Enhanced Communications with SARSA Agent}

The parameters and initial conditions used in this scenario of the the proof-of-concept are the following:

\begin{itemize}
    \item SARSA Agent configuration parameters:
    \begin{itemize}
        \item Epsilon-greedy policy $\epsilon$ of 2$\%$;
        \item Learning rate $\alpha$ of 20$\%$; and
        \item Discount factor $\gamma$ of 80$\%$
    \end{itemize}
    \item Pub/Sub queue threshold limit (triggers agent action) = 50\%
    \item Agent actions: bandwidth increased or reduced by 10\%
    \item Maximum number of attempts = 400
\end{itemize}

The SARSA configuration parameters are fixed for all evaluations in this scenario 2. Typical values were used and the impact of the variation of the values of these parameters was not considered in this evaluation and left for future works.

In terms of the PSIoTRL operation and SARSA evaluation, the agent action in the orchestrator is triggered by the aggregator anytime one of its queues occupation goes beyond
the configured threshold limit (Figure \ref{fig:PSIoTRL-Communication}).

The initial condition with SARSA are the same as those applied for the scenario 1 as indicated in Table \ref{TableProofofConcept}. Ten runs were executed for each of the four initial states and the obtained results are summarized in Table \ref{ResultCase2} with a confidence interval of 5\%.

\begin{table}[!htp]
\centering
\caption{Queue Occupation, Packet Loss and Link Occupation with SARSA.
\label{ResultCase2}}
\resizebox{\textwidth}{!}{%
\tiny{
\begin{tabular}{ccccc}
\hline
\textbf{\begin{tabular}[c]{@{}c@{}}Initial QS \\ (B1i,B2i,B3i)\end{tabular}} & \textbf{\begin{tabular}[c]{@{}c@{}} Final QS\\ (B1,B2,B3)\end{tabular}} & \textbf{\begin{tabular}[c]{@{}c@{}} Final TR\\ (T1,T2,T3)\end{tabular}} & \textbf{\begin{tabular}[c]{@{}c@{}} Packet Loss \\ (P1,P2,P3)\end{tabular}} & \textbf{ Link Occupation} \\ \hline
AL,BL,BL & (48,12,22) & (49,35,13) & (0,0,0) & 97 \\
AL,AL,BL & (47,47,23) & (50,35,12) & (0,0,0) & 97 \\
AL,AL,AL & (48,48,87) & (49,35,14) & (0,0,0) & 98 \\
+AL,BL,BL & (24,24,26) & (63,20,11) & (28,0,0) & 94 \\ \hline
\end{tabular}%
}}
\begin{itemize}
\item QS: Queue State; TR: Transmission Rate; AL: 80$\%$; BL: 20$\%$; +AL: 120$\%$.
\end{itemize}
\end{table}

\subsection{Bandwidth Allocation with SARSA for Buffer Flushing  - Results}

First of all, it is essential to highlight that this analysis's objective is to verify that the
SARSA algorithm enhances bandwidth allocation for an IoT aggregator using specifically a Pub/Sub method to exchange data. In this context, enhanced bandwidth allocation means that the constrained bandwidth resources can be tuned and, consequently, better used for attained QoS performance parameters like delay, jitter, and packet loss.

The evaluated parameters are queue occupation, link occupation and packet loss.

The algorithm's ability to bring the queue state back to a threshold limit is not an effective performance parameter but demonstrates that the algorithm enhances the usage of the constrained bandwidth resources. For example, keeping the occupation of queues 1 and 2 below 50\% would imply having the delay and jitter for the IoT data belonging to the specific Pub/Sub topics within its application's requirements.

The link occupation performance parameter evaluated in the runs indicates the efficient use or not of the bandwidth available, and the packet loss performance parameter is relevant to Pub/Sub topics sensitive to data loss.

\begin{figure}[!htp]
\centerline{\includegraphics[scale=0.65]{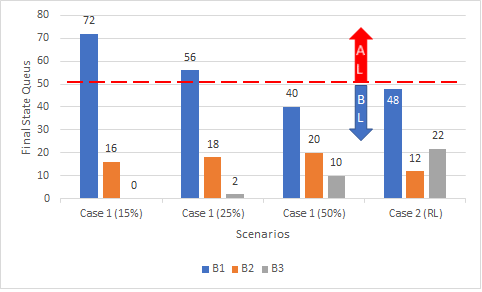}}
\caption{Final Queue Occupation - Initial State = $AL,BL,BL$.\label{fig:ALBLBL}}
\end{figure}

\begin{figure}[!htp]
\centerline{\includegraphics[scale=0.65]{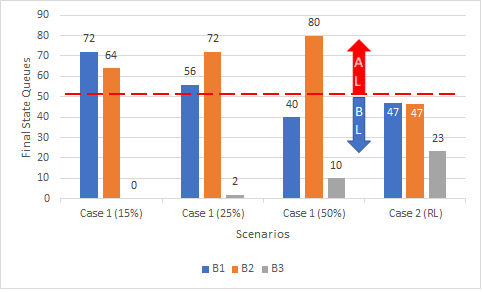}}
\caption{Final Queue Occupation - Initial State = $AL,AL,BL$.\label{fig:ALALBL}}
\end{figure}

\begin{figure}[!htp]
\centerline{\includegraphics[scale=0.65]{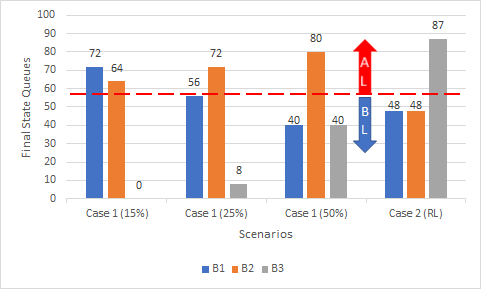}}
\caption{Final Queue Occupation - Initial State = $AL,AL,AL$.\label{fig:ALALAL}}
\end{figure}

\begin{figure}[!htp]
\centerline{\includegraphics[scale=0.65]{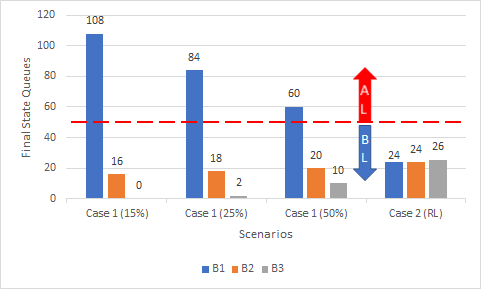}}
\caption{Final Queue Occupation - Initial State = $+AL,BL,BL$.\label{fig:+ALBLBL}}
\end{figure}

The Figures \ref{fig:ALBLBL}, \ref{fig:ALALBL}, \ref{fig:ALALAL} and \ref{fig:+ALBLBL} present the final buffer occupation with bandwidth allocation for buffer flushing done by the orchestrator using fixed rule (case 1) and SARSA (case 2).

For all initial states $(AL, BL, BL;~AL,AL,BL;~AL,AL,AL~and~+AL,BL,BL)$ bandwidth allocation with SARSA performed better that the fixed rule using increments of 15\%, 25\% and 50\% in the buffer flushing bandwidth. SARSA algorithm brought all buffers to the final condition $BL,BL,BL$.

In Figure \ref{fig:ALALBL}, the case 1 (50$\%$) was the unique fixed rule option that succeeded to bring one of the queue occupation (queue 3) below the threshold limit.

In Figure \ref{fig:ALALAL}, the bandwidth allocation with SARSA performed again better than the
fixed rule algorithm. However, while the two priority buffers $B1$ and $B2$ are below the limit, buffer $B3$ (least effort) was set above the limit due to its lower priority.

Figure \ref{fig:+ALBLBL} corresponds to the evaluation scenario in which we want to observe the behavior of the fixed rule bandwidth allocation and SARSA when buffer $B1$ is already experiencing
packet loss (above 100\% capacity). Figure \ref{fig:+ALBLBL} shows that the SARSA algorithm is again the most efficient approach to bring all buffers below the threshold limit.

\subsection{Allocated Bandwidth per Queue, Link Occupation and Packet Loss}

In this experimentation, the objective is to use the minimum amount of link bandwidth per buffer to transmit data. Buffer priority should be preserved, and the aimed final condition is to bring all buffers to the $BL$ state after bandwidth allocation.

\begin{figure}[!htp]
\centerline{\includegraphics[scale=0.65]{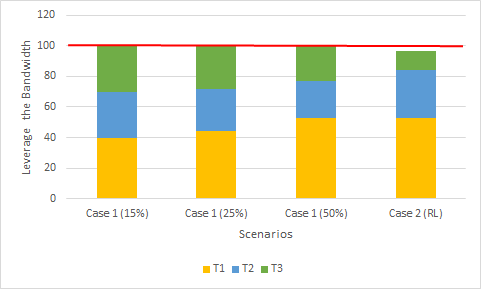}}
\caption{Allocated Bandwidth per Queue and Link Occupation.\label{fig:Bandwidth}}
\end{figure}

Figure \ref{fig:Bandwidth} illustrates two related aspects of link occupation. Firstly, it shows that the SARSA allocation of bandwidth per buffer is aligned with buffer priorities. In fact, more
bandwidth is allocated to $B1$ than to $B2$, and, in turn, $B2$ gets more bandwidth than $B3$. This
result is fully consistent with the defined buffer priorities. The fixed rule method allocates
bandwidth to buffer $B1$ and splits the remaining bandwidth among the other buffers.

A second result illustrated in Figure \ref{fig:Bandwidth} is link occupation. The result indicates that the SARSA algorithm is more efficient than the fixed rule allocation method. SARSA succeeds
in bringing all buffer occupation levels bellow the defined threshold and, concomitantly, preserves link utilization below 100\%.

Packet loss is indicated in Tables \ref{Result15Case1}, \ref{Result25Case1}, \ref{Result50Case1} and \ref{ResultCase2}. The allocation of chunks of bandwidth for IoT data flushing in queues is done proactively once the 50\% buffer occupancy threshold is reached. Consequently, no packet loss occurs for three out of four cases considered in the PSIoTRL evaluation $(AL, BL, BL;~AL, AL, BL~and~AL, AL, AL)$. It happens that the buffer flushing bandwidth is adjusted before any packet loss could occur.

The case $+AL, BL, BL$ starts with buffer overloaded and, as such, packet loss does occur since either the fixed rule method or the SARSA algorithm takes some time to allocate bandwidth to the overloaded buffer and fix the problem. As expected, SARSA takes more time to allocate bandwidth due to the learning process inherent to the algorithm and, consequently, presents a more significant packet loss.

\section{Final Considerations} \label{sec:conclusion}

This work has presented a solution to address the problem of network resources allocation in the context of massive IoT data distribution. It has presented a PSIoTRL framework that introduces an architecture that aim to manage the allocation of limited network resources to aggregators.

As highlighted in the simulations, the SARSA agent in PSIoTRL was able to reconfigure the queues' flushing transmission rate efficiently, providing in 100$\%$ of the tested cases, that these queues reached the BL (below threshold limit) final condition with less link bandwidth usage. The PSIoTRL demonstrates that a solution based on SARSA is an efficient reinforcement learning approach for enhancing resource (bandwidth) allocation in the Pub/Sub-based massive IoT data exchange context.

Future works will address SARSA computation scalability concerning the granularity for allocating new chunks of bandwidth to flush IoT data and adjust the Pub/Sub queue occupancy. The impact of the SARSA configuration parameters (epsilon-greedy policy, learning rate, and discount factor), in the algorithm time response, with a focus on the learning rate, will also be evaluated.

\section*{Acknowledgments}
Authors want to thanks CAPES (Coordination for the Improvement of Higher Education Personnel) for the master's scholarship support granted.

%
%

\bibliographystyle{splncs04}
\bibliography{allbib}

\end{document}